\documentclass[11pt,a4paper]{article}
\usepackage{times,latexsym}
\usepackage[T1]{fontenc}
\usepackage{graphicx}
\usepackage{caption}
\usepackage{amsmath}
\usepackage{amssymb}
\usepackage{pgfplots}
\pgfplotsset{compat=newest}
\usepackage{comment}
\usepackage{enumitem,kantlipsum}
\usepackage{times}
\usepackage{booktabs}
\usepackage{array,multirow}
\usepackage{hyperref}
\usepackage{colortbl}
\usepackage{arydshln}
\usepackage{pbox}
\usepackage{url}

\usepackage{siunitx}

\usepackage[acceptedWithA]{tacl2018v2}

\usepackage{xspace,mfirstuc,tabulary}

\newif\iftaclinstructions
\taclinstructionsfalse 
\iftaclinstructions
\renewcommand{\confidential}{}
\renewcommand{\anonsubtext}{(No author info supplied here, for consistency with TACL-submission anonymization requirements)}
\newcommand{\instr}
\fi

\iftaclpubformat 

\else

\fi

\newcommand{\ella}[1]{{\color{blue}{#1}}}

\newcommand{\tocheck}[1]{{#1}} 
\newcommand{\overlap}[1]{\textcolor{magenta}{#1}} 
\newcommand{\ellanew}[1]{{#1}} 
\newcommand{\ssnew}[1]{{#1}} 
\newcommand{\ssnewer}[1]{{#1}} 
\newcommand{\tochecknew}[1]{\textcolor{black}{#1}}

\newcommand{\Section}[1]{\S\ref{#1}}
\newcommand{\Table}[1]{Table~\ref{#1}}
\newcommand{\Figure}[1]{Figure~\ref{#1}}
  
\title{Quantifying Cognitive Factors in Lexical Decline}

\author{
	David Francis$^{1}$ \hspace{2.5cm}
	Ella Rabinovich$^{1}$ \hspace{2.5cm}
	Farhan Samir$^{1}$ \\
	\textbf{David Mortensen}$^{2}$ \hspace{3cm}
	\textbf{Suzanne Stevenson}$^{1}$ 
	\vspace{0.1cm} \\
	$^{1}$Department of Computer Science, University of Toronto, Canada \\
	$^{2}$Language Technologies Institute, Carnegie Mellon University, USA
	\vspace{0.1cm} \\
	\texttt{\{dfrancis, ella, fsamir, suzanne\}@cs.toronto.edu} \\
	\texttt{dmortens@cs.cmu.edu}
}

\date{}

\begin{document}
\maketitle

\begin{abstract}
We adopt an evolutionary view on language change in which cognitive factors (in addition to social ones) affect the fitness of words and their success in the linguistic ecosystem.  
Specifically, we propose a variety of psycholinguistic factors --- semantic, distributional, and phonological --- that we hypothesize are predictive of lexical decline, in which words greatly decrease in frequency over time. Using historical data across three languages (English, French, and German), we find that most of our proposed factors show a significant difference in the expected direction between each curated set of declining words and their matched stable words.  Moreover, logistic regression analyses show that semantic and distributional factors are significant in predicting declining words. Further diachronic analysis reveals that declining words tend to decrease in the diversity of their lexical contexts over time, gradually narrowing their `ecological niches'.

\end{abstract}

\section{Introduction}
\label{sec:introduction}


Many researchers, from \citet{schleicher1863darwinsche} up to the present \cite{croft2000explaining,oudeyer2007language,atkinson2008languages,thanukos2008look, turney2019natural}, have drawn analogies between biological evolution and the evolution of languages --- their structure, their semantics, and their lexicons. Lexically speaking, as Schleicher first pointed out, diachrony can be viewed as a struggle for survival by individual words whose propagation into future generations is contingent on their continued 
fitness for one or more niches in the ecology of the speech community --- as determined by a host of factors.
Here we study the question of \textit{lexical decline} --- a gradual decrease in frequency and ultimate obsolescence of words.

\begin{figure}[h!]
\centering
\resizebox{\columnwidth}{!}{
\includegraphics{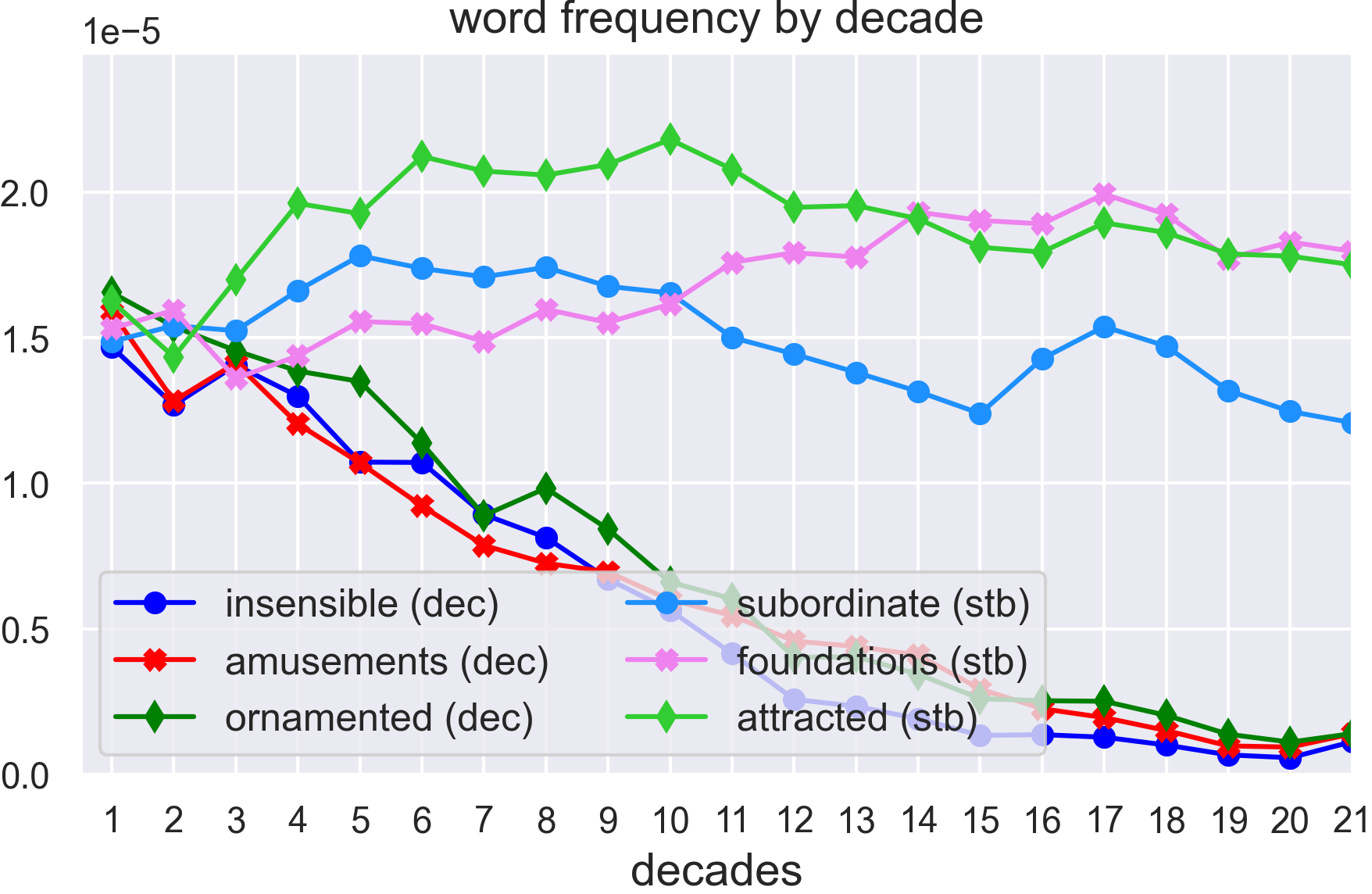}
}
\caption{Matched dec:stb word pairs illustrated by dark (dec) and light (stb) shade of the same color. 
}
\label{fig:freq-by-decade}
\end{figure}

What explains that the word `amusements' has declined in the last 200 years, but `foundations' has not, as shown in \Figure{fig:freq-by-decade} (along with other similar pairs)? Social factors clearly play a role, as changes in culture and technology may lead words to fall in and out of use.  But cognitive and linguistic factors also influence lexical survival (e.g., \citet{vejdemo2016semantic}). 
Words that are semantically similar to many other words may come to be used less because of intense competition in the cognitive process of lexical access \cite{chen2012competition}. 
Words that can occupy many niches, distributionally speaking, should have better chances of being learned and used, and therefore perpetuated, than words that are confined to a narrow range of contexts or senses \cite{altmann2011niche,stewart2017making}. On the other hand, words that are phonologically very different from other words may suffer because they are more difficult to access, sitting as they do at the formal fringes of the mental lexicon \cite{edwards2004interaction}. 
In other words, we suggest that semantic, distributional, and phonological  factors all play a role in the natural selection of words.

While attention to predicting what words will emerge, live, and die goes back to Schleicher, there is relatively little computational work on this subject (examples include \citet{cook2010automatically,hamilton2016diachronic,xupredictability,ryskina2020new}). In particular, little attention has been paid to the factors that contribute to 
lexical decline (but see \citet{vejdemo2016semantic} for related work on lexical replacement).
This is unfortunate because understanding this phenomenon answers an important scientific question about language change --- how lexicons become as they are. We ground these phenomena in an evolutionary model of linguistic diachrony in which fitness is influenced by independently motivated cognitive processes like lexical access.

Our study spans 20 decades and three languages. We find that there are consistent factors --- semantic, distributional, and phonological --- that predict whether a word is likely to substantially decline in frequency. We propose that our observations are consistent with a model where there is a feedback loop between cognition and usage driving the diachronic development of lexicons.\footnote{All data and code is available at \url{https://github.com/ellarabi/linguistic_decline}.}


\section{Related Work}
\label{sec:related-work}
There is a vast body of research on lexical change of various kinds; in this section we focus on work involving the birth and death of words, as it is most closely related to our study here.

Lexical neology --- introduction of new words --- is one of the most evident types of lexical change.  Various computational studies have suggested a range of factors underlying the phenomenon of neology, including semantic, distributional, and phonological influences.
\citet{ryskina2020new} show that lexical neology can be partly explained by the factor of \textit{supply} --- new words tend to 
emerge in areas of semantic space where they are needed most, i.e., areas exhibiting relative sparsity. 
Drawing on theories of patterns of word growth \citep{metcalf2004predicting, cook2010automatically, chesley2010predicting}, additional studies suggest that (among other factors) greater linguistic distribution across individuals and topics plays a significant positive role in the fate of novel lexical items in online forums \citep{altmann2011niche, stewart2017making}.
Considering phonological factors, \citet{xupredictability} show that new words emerge under the joint constraints of \textit{predictability} and \textit{distinctiveness}: they efficiently recombine elements from existing word forms, yet are sufficiently distinctive to reduce confusion.
Viewing the lexicon as an evolving eco-system, with interacting birth and death of words, we hypothesize that analogous factors will play a role in lexical decline as in neology.

Compared to research on 
neology, the work on lexical obsolescence and loss is relatively sparse. 
While the study of neology often draws on occurrence of new word forms in contemporary corpora, 
in contrast, the study of lexical loss inherently relies on the availability (and the quality) of large diachronic textual resources. \citet{tichy2018lexical} proposes a methodology for identifying declining words in such a corpus, and performs qualitative analysis of a sample of such words, focusing on spelling standardization and changes in word-formation strategies. 
Using the Google-books dataset \citep{michel2011quantitative}, \citet{petersen2012statistical} 
study the `death rate' of words primarily stemming from misspellings and print errors typical to historical corpora, focusing on the rate and not the causes of linguistic decline.

Other work touches on lexical decline less directly, but explores potential predictive factors (as we do) in related processes --- factors that may also play a role in decline. \citet{hamilton2016diachronic} consider the factors that influence meaning shift --- rise and decline of meanings \textit{within} a word (rather than of words themselves) --- and find that both word frequency and number of meanings play a 
role.
\citet{turney2019natural} track the evolution of $4$K English synsets, attempting to predict a synset `leader' --- the member of the synset with highest frequency.
They find the current `leadership' of a word to be the most predictive factor of its future status as a `leader', again illustrating the driving force of frequency in lexical status. However, while a word may become a synset leader at the expense of other words, this work does not perform a systematic study of factors predictive of lexical decline.

Finally, \citet{vejdemo2016semantic} conducted a study of lexical replacement --- a closely-related but narrower phenomenon than lexical decline --- exploring similar semantic factors to those we investigate here using a markedly different methodology.  Their study is focused on a small set of core vocabulary in Indo-European languages (``Swadesh list'' words, \citet{swadesh1952lexicostatistic}, from \citet{pagel2007}).  Our research here addresses a much broader phenomenon of general lexical decline, and proposes a wider range of factors influencing that process.

\section{Overview of Our Approach}
\label{sec:overview}

Motivated by the perspective of the lexicon as an evolving ecosystem, in which words are subject to various cognitive pressures that can influence their `survival', we aim here to identify factors that may be indicative of words that are likely to decline.  Specifically, we propose factors that, when calculated at a given time in history, $t$ (in our study, $1800$--$1810$), are hypothesized to be predictive of lexical decline during a subsequent stretch of time, up to $t{+}n$ (here $2000$--$2010$).

\ssnew{First we note one potential factor
whose influence on decline is \textit{not} explored here: that of a word's frequency.  Having seen that (relatively higher) frequency of a word is the single best predictor of future (relatively higher) frequency \citep{turney2019natural}, a natural hypothesis is that lower frequency may conversely be predictive of future decline.  However, since relatively low frequency may indicate a word already `on its way out', we instead control for frequency: Given words of similar frequency, we explore what other properties are most predictive of which will \textit{subsequently} decline and which survive.}

As noted in \Section{sec:introduction}, we consider that semantic, distributional, and phonological factors all may play a role in lexical decline, due to their influence on the ease or difficulty of learning and accessing of words, which may impact a word's continued role in the lexical ecosystem.  
\ssnew{Here we provide the motivation for the factors we consider; \Section{sec:factors-mm} provides detail on how they are computed.  While this discussion may suggest causal relationships (e.g., words decline because their lexical access is more difficult), our subsequent analyses focus on correlations of the factors with decline, and are thus agnostic with respect to causality.}

We consider several semantic factors, drawing on inspiration from the acquisition and processing literatures.  First, we consider the role of the semantic space a word occurs in.  While some work has found that lexical access is facilitated by having dense semantic neighborhoods (having many closely related words) \citep{buchanan2001}, other work has noted their inhibitory effect on semantic processing \citep{mirman2008attractor}, in line with findings of inhibitory competition in phonological neighborhoods \cite{marslen-wilson1990activation, dahan2001subcategorial}.  Such inhibitory effects may underlie the observation that words in semantically dense (i.e., more competitive) environments are more likely to be driven out, to the benefit of others that can potentially be used to express roughly the same meaning (e.g., \citet{breal1904essai, vejdemo2016semantic}). Thus, similarly to \citet{ryskina2020new}, 
we estimate the density of a word's immediate semantic neighbourhood, where we predict words with a higher \textbf{semantic density} to be more likely to decline.

Next, we consider properties of the semantics of the word itself. 
Psycholinguistic studies have found that more concrete words --- roughly, those referring to a perceptible entity --- are learned and retrieved more easily (e.g., \citet{james1975role,degroot2000hard}).
Moreover, concrete words may form a more stable subset of the lexicon 
(\citet{swadesh1971origin}; 
cf.~a similar finding in \citet{vejdemo2016semantic} using imageability ratings, a notion that is highly correlated with concreteness).
Because words conveying a more concrete meaning appear more likely to survive,
we consider the level of \textbf{concreteness} of a word as a second semantic factor, where lower concreteness predicts a higher chance of decline. 

Additionally, having a higher degree of polysemy 
has been shown to have a facilitatory effect on a word's lexical access, due to multiple related senses contributing to aggregate activation of the word \citep{jastrzembski1981multiple, rodd2002making}.  Access to a word across many senses may similarly lead to greater survivability \citep{vejdemo2016semantic}, and we thus predict that words with a higher \textbf{number of meanings}
will be less likely to fall out of use in a language.

In addition to the influence of semantic properties, others have proposed a central role for distributional factors in lexical learning and processing \citep{mcdonald2001rethinking,jones2017context}.  In particular, words that occur in more varied contexts are easier both to learn \citep{johns2016} and to access \citep{mcdonald2001rethinking}.  In addition, words with broader topical dissemination tend to become more robustly entrenched into the lexicon \citep{ altmann2011niche, stewart2017making}; conversely, we expect that words that occur in a narrower range of contexts will be more apt to fall out of use.  This too follows our lexical evolution perspective: just as species that can occupy many niches in a natural ecology are more likely to survive,  generation to generation, lexemes that occupy many niches in the linguistic ecology are less likely to face extinction (or decline).
We adopt the distributional factor of \textbf{contextual diversity} to model this fact.

Like semantic and distributional effects, phonological effects are also known to interact, in a complex way, with lexical processing, and we hypothesize that such factors may also be predictive of 
lexical decline. 
For example, psycholinguistic studies have found that phonotactically typical words are recognized more quickly 
than atypical words \cite{vitevitch1999phonotactics}. We correspondingly predict that \textbf{phonological typicality} will be associated with lower rates of lexical decline.\footnote{We also considered orthographic typicality; this measure
correlated highly with phonological typicality ($r$ of over $0.6$ in all $3$ languages), and showed precisely the same pattern as phonological typicality across declining and stable words.}

As with semantic neighborhoods, psycholinguistic experiments have also found mixed effects of phonological neighborhoods on lexical processing: both competition among similar phonological forms (as noted above, \citet{marslen-wilson1990activation, dahan2001subcategorial}), as well as potential facilitation from having a higher number of phonologically-close neighbors
\cite{yates2004influence,vitevitch2002influence,marian2006phonological}.
Given a preponderance of evidence of facilitatory effects on lexical processing, we predict that \textbf{phonological density} will be inversely correlated with lexical decline.\footnote{\ssnewer{The mixed effects of competition and facilitation within a phonological neighborhood may help explain why, as noted earlier, new word forms tend to show a tension between predictability and distinctiveness \cite{xupredictability}.  Here we predict an inverse correlation of phonological density and lexical decline, but future research on the role of neighborhoods in lexical access will be necessary to reconcile these viewpoints.}}

Finally, we predict that words with greater \textbf{phonological complexity} 
--- for our purposes, longer in terms of the number of syllables --- will be more likely to decline. This hypothesis follows from the speculation that words are processed as sequences of syllables rather than sequences of phonemes, and that longer words are more effortful to process. Specifically, we hypothesize that words with higher number of syllables (per phoneme) will be more likely to decline.

Table~\ref{tab:factors-and-predictions} summarizes the seven proposed factors, grouped by categories, as well as their predicted direction of correlation with the tendency of a word to decline.
Note that none of these factors operates in isolation, and they may interact to push in the same or different directions; for example, a word with competition from many semantically similar alternatives may also be highly phonologically typical or simple.  
%
%
To be clear, we are not claiming that these are the only factors predictive of the decline of words.  \ssnew{For example, other linguistic factors, such as pragmatic influences, are likely involved, but we limit our study to lexical properties that are readily extractable from the available historical resources such as corpora and dictionaries.  Moreover, s}uch cognitive factors necessarily interact with extensive sociological and cultural trends that impact word usage (e.g., the decline in systems of aristocracy, or a shift in medical terminology).  Here we explore whether internal cognitive factors may play a role beyond these broad extra-linguistic influences.

\begin{table}[ht!b]
  \small
  \centering
  \resizebox{\columnwidth}{!}{  
  \begin{tabular}{llc}
    \toprule
     &  & Predicted Corr.\\
    Group & Factor & w/Decline\\
    \midrule
          & semantic density & $+$ \\
    \bfseries semantic & concreteness & $-$ \\
          & number of meanings & $-$ \\
    \midrule
    \bfseries distributional & contextual diversity & $-$ \\
    \midrule
          & phon typicality & $-$ \\
    \bfseries phonological & phon density & $-$ \\
          & phon complexity & $+$ \\
    \bottomrule
  \end{tabular}
  }
  \caption{Factors and their predicted correlation, positive ($+$) or negative ($-$), with decline. 
  }
  \label{tab:factors-and-predictions}
\end{table}

In order to assess the factors both individually and as a collection, we perform two kinds of analyses.  We identify a set of words that decline in usage over a $200$-year period, and pair those with a set of 
words that are stable in frequency over the same period. 
We first consider whether the values (in the initial decade) of each of these proposed factors differs in the expected direction between the declining and stable words.  Next, we 
see which factors may be most explanatory of decline when 
the set of $7$ factors are used collectively in a logistic regression analysis.
In \Section{sec:m-and-m} we describe how we select our declining and stable words, and estimate the above factors, and in \Section{sec:results} we present the results of these two analyses.  We follow this in \Section{sec:diachronic-analysis} with further diachronic analysis of the pattern of contextual usage in how words decline. 

\section{Materials and Methods}
\label{sec:m-and-m}

\ssnewer{Our goal is to explore whether the factors identified above can indeed distinguish words at a time $t$ that will decline over a subsequent period of time $t+n$, from words that remain relatively stable over that same time period.  To this end, we develop measures to identify a set of words that have declined over a historical period, and a set of stable words for comparison.  However, we cannot form our experimental word sets by simply selecting words randomly from each of these lists.  Because confounding lexical properties (such as frequency) may interact with our identified factors of interest, we must adopt a more controlled approach, standard in cognitive research, of matching our declining and stable words on a set of potential confounds.  In \Section{sec:motiv-matching}, we first motivate our approach to forming our experimental items -- pairs of declining and stable words matched on covariate properties.  We then detail how those word pairs are selected (\Section{sec:word-sampling}), and finally explain how we estimate our identified factors of interest over these experimental items (\Section{sec:factors-mm}).}

\subsection{Motivation for matching pairs of declining and stable words}
\label{sec:motiv-matching}

\ssnewer{As noted earlier, frequency at time $t$ (the start time of our analysis) may be a powerful indicator of which words are already in the process of decline.  Indeed, we found random samples of stable words to be on average 2--3 times more frequent than declining words (with stable and declining measured as in \Section{sec:word-sampling}) at the initial time $t$.  Initial frequency is thus a confounding factor on which the declining and stable words need to be matched. Word length is another potential confound we noted: in addition to being highly correlated with frequency \citep{Zipf1936}, word length may mask (or otherwise interact with) the factors we have identified as related to decline.  For example, shorter, more frequent words tend to have more meanings as well.  While it may be of some limited interest to show that stable words tend to be shorter than declining words, we were interested to see the effect of our richer lexical factors beyond this. Finally, we suspect that words with different parts of speech show different patterns of decline; therefore, we also controlled for this potential confound.}

\ssnewer{One possibility would be to ``range-match’’ the overall sets of declining and stable words on these covariates --- i.e., picking words in the same frequency and length ranges, and with an overall similar distribution of POS.  However, this approach is not sufficient, since these covariates can interact with our factors of interest. }
\tocheck{For example,}
\ssnewer{the number of meanings of words correlates with frequency \citep{Zipf1949}. While there may be differences in polysemy of words at the same frequency that are predictive of decline, when compared over a broad range of frequencies, the differences in numbers of meanings across that range may swamp out differences in stable and declining words of a particular frequency. }
\tocheck{Detecting such differences may require complex statistical models with many parameters to capture this kind of interaction between our factors of interest and the confounding variables. } 

\ssnewer{To address this, we take a simpler and more controlled approach, standard in human experimental work, of pairing each declining word with a stable word with matching values on these three covariates.  That is, for each declining word, we find the most stable word (above a certain stability threshold) of the same POS, such that each pair has a very close value of frequency and word length (as detailed below).  Because our resulting experimental items are words pairs, we then perform pairwise statistical analyses to see whether declining and stable word pairs matched on these key covariates display the predicted difference in each of the factors we explore.  (Note that controlling the covariates across the declining and stable words yields declining and stable word sets that are not statistically independent, such that pairwise statistical analyses are recommended.)}

\subsection{Selecting Declining and Stable Words}
\label{sec:word-sampling}

We select two sets of words, in each of English, French, and German, to be used for testing our hypotheses on factors that affect lexical decline: (1) words that gradually declined in their frequency from $1800$ to $2010$, 
and (2) control words 
that maintained a relatively stable frequency across the $21$ decades.
The words were selected from the Google ngrams dataset \citep{michel2011quantitative}, where individual years (and, consequently, yearly word frequencies) were accumulated into decades, the time unit of our analysis.

\subsubsection{Identifying a Set of Declining Words}
\label{sec:dec-words}

We aim for the declining set to contain words that were in common use during the first decade of the $19$th century ($1800$--$1810$), but gradually have become much less common in contemporary language.\footnote{We exclude words that underwent orthographic change, but preserved meaning and phonetic form, from this study.} We define a declining word as one exhibiting a period of gradual, steady decline (to very low, possibly $0$, frequency), followed by a period of infrequent usage (at or near $0$).

To select such words, following \citet{stewart2017making}, we define a model based on piece-wise linear regression fitting the frequencies of a word during the $21$ decades. Formally, we find the curve of the following form that has the least mean-squared error (MSE) to the word's frequency curve:

\[
x(t) = \begin{cases}
a (b - t) \hspace{0.4cm} \mathrm{if} \; t \leq b\\
0 \hspace{1.5cm} \mathrm{if} \; t > b 
\end{cases}
\]

\noindent 
where $t$ is time in decades, and $a$ and $b$ are parameters defining the curve: both $a$ and $b$ are positive, and $b$ is the value within the $(0\mbox{-}21)$ range of decades that minimizes the MSE. We thus fit the word's frequencies to a curve with a declining piece (crossing the x-axis of $0$ frequency at $b$), and a `zero' piece (horizontal at frequency $0$).

We define the decline metric as the MSE between these two pieces of the fit curve and the true frequencies.  \ssnewer{This MSE metric ensures that words are ranked highly if they show consistent temporal decline, followed by a period of stable usage near $0$ -- the target behavior for words to be considered as having declined.}
\ssnew{We normalize the frequencies of each word across the $21$ decades because we are interested in the relative amount of change in that word's frequency over time.  Having observed that words with higher average frequency generally yielded higher MSE, this normalization adjusts to put words at different frequencies on a level playing field in calculating the MSE.  (See Appendix A.1 for further detail and illustration of this normalization step.)}

Words ranked highest according to the defined metric were considered as declining candidates, and were subject to further automatic filtering to ensure their suitability for our analysis; e.g., we excluded words shorter than $4$ characters or whose relative frequency was less than $5{\times} 10^{-6}$ in the first decade of the $19$th century, 
or whose piece-wise regression crossed the x-axis within less than $10$ decades from the starting point.\footnote{The latter condition removed OCR errors, such as \textit{fome} for \textit{some}, that are more evident in earlier decades.} 

Additional manual filtering was then performed by native speakers of English, French, and German with a linguistics background.  This inspection aimed at excluding multiple forms (e.g., inflections) of the same word, since our predictors 
are likely to have a similar effect on all words stemming from the same lemma.  We replaced multiple variants of a word (such as German `ansehnliche', `ansehnlich', and `ansehnlichen') with a single representative that had the  highest frequency among them in $1800$--$1810$ (in this case, `ansehnliche').\footnote{\ssnew{We select a single representative word (rather than, e.g., averaging our predictors over multiple alternatives) to facilitate pairwise word matching, since frequency, length, and POS can differ across a set of morphologically related words.}}

Our final sets of declining words comprise $300$ words each for English and French, and $250$ words for German, due to the relative sparsity of the latter in the historical part of the corpus.

\subsubsection{\ssnewer{Identifying the Matched Stable Words}}

\ssnewer{As motivated in \Section{sec:motiv-matching}, we next select a matched stable word for each declining word in our datasets.}
Specifically, we match each declining word with a stable counterpart that maintained relatively constant frequency over the period of $1800$--$2010$. The `stability' criterion was measured by the MSE of a word's true frequencies to the \textit{horizontal} trend of best fit (using the same normalization of frequencies as for declining words; see Appendix A.1).

The matching procedure paired each declining word with a stable counterpart, ensuring similarity in three properties that could introduce bias into the analysis:
the initial frequency of a word ($\pm10\%$), its length in characters ($\pm2$ characters, with the additional restriction that the sum of lengths of all stable words must be within $1$ of the sum of lengths of all declining words), and its POS (i.e., nouns were matched with nouns, adjectives with adjectives, etc.); \ssnew{see Appendix A.2.}  For example, 
Figure~\ref{fig:freq-by-decade} in \Section{sec:introduction} illustrates the diachronic trends of three matched English word-pairs that have various initial frequency, POS, and length.

\ssnew{The carefully curated sets of declining and stable words facilitate rigorous analysis of the factors that we hypothesize are predictive of lexical decline.  Specifically, the matched sets enable comparison of the factor values between pairs of words --- one stable and one declining --- that are matched on key linguistic properties at the starting point of our analysis. In this way, we control for these matched linguistic properties, and see how differences in our identified factors correlate with the final fate of the words --- gradually experiencing lexical decline, or soundly persisting across $210$ years of language use.} \ssnew{Appendix A.3 provides examples of matched word-pairs for the three languages -- English, French and German.}


\subsection{Estimating Factors Predictive of Decline}
\label{sec:factors-mm}

Here we describe how we estimate each of the $7$ features we hypothesize are predictive of lexical decline, in each of the $3$ languages; cf.~\Table{tab:factors-and-predictions}. In each case, we calculate the feature based on its value at the beginning of the time period we consider ($1800$--$1810$), except as noted below.

\begin{figure}[hbt]
\centering
\resizebox{\columnwidth}{!}{
\includegraphics{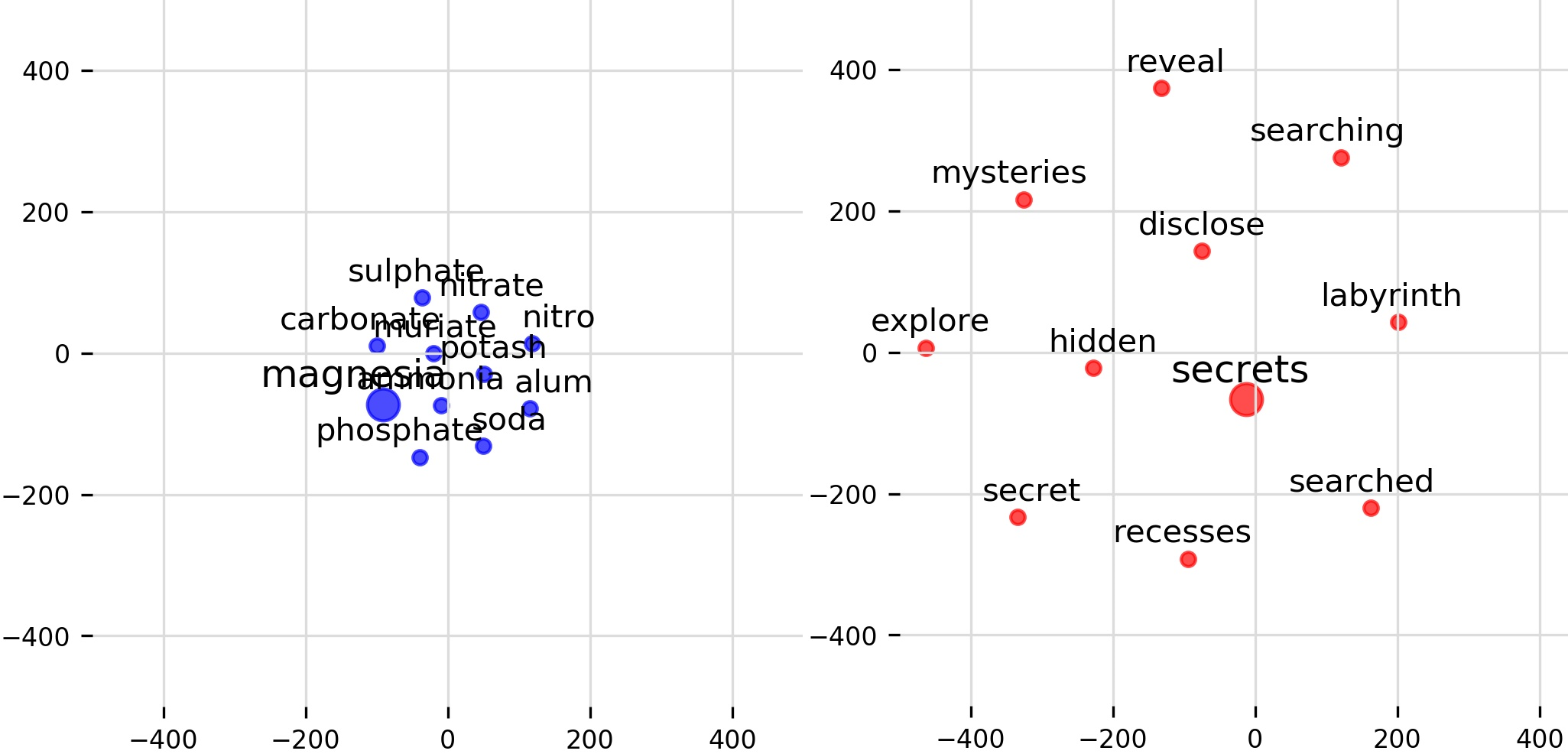}
}
\caption{\ellanew{t-SNE projection of the 10-closest neighbors in the semantic space of the matched words `magnesia' (dec):`secrets' (stb). The semantic neighborhood of `magnesia' (left) is denser, compared to that of `secrets' (right). \texttt{SemDens} values for these words are $0.872$ and $0.500$, respectively.}}
\label{fig:sem-neighborhood}
\end{figure}

\paragraph{Semantic Density (\texttt{SemDens}).}
\label{sec:semantic-density-mm}
We define semantic density as the average similarity of a word to its $10$ nearest neighbors in semantic space.\footnote{Using $20$ or $50$ neighbors gave similar results; \ssnew{Pearson correlations between \texttt{SemDens} using $10$ neighbors and \texttt{SemDens} using $20$ or $50$ neighbors both yield $r=0.99$.}} We use the historical embeddings made available by \citet{hamilton2016diachronic},\footnote{\ssnew{We use the word2vec (SGNS) versions \citep{mikolov2013distributed}, from \url{https://nlp.stanford.edu/projects/histwords/}.}}
and use cosine similarity between two representations in the semantic space. 
The three languages vary in availability of these semantic representations. English  benefits from ample historical data, and all $600$ words were found.  For French, $530$ out of $600$ word representations were found (balanced between stable and declining sets); we interpolated semantic density values for the missing words by \ssnew{using one of the most popular data imputation methods} --- assigning them the mean \texttt{SemDens} value of the $530$ available representations. We exclude German from analysis of this factor because only $22$ of our German declining and stable words have historical embeddings.
Figure~\ref{fig:sem-neighborhood} illustrates the prediction that a denser semantic neighborhood is observed for a declining word (here `magnesia', left) compared to its corresponding stable word (here `secrets', right).

\paragraph{Concreteness (\texttt{Conc}).}
\label{sec:distinctiveness}
\citet{snefjella2019historical} released a dataset of (automatically inferred) historical by-decade concreteness ratings for over $20$K English words, dating back to $1850$. Assuming that the concreteness of  individual words did not undergo a significant change during the period $1800$--$1850$, we use the scores from $1850$ as a close approximation of English concreteness ratings in $1800$--$1810$.
With no access to historical concreteness norms for French and German, we only calculate this feature for English.
Because only $461$ out of our $600$ English words have a concreteness rating in the \citet{snefjella2019historical} dataset, we use an adaptation of the approach by \citet{tsvetkov2013cross} to infer concreteness values for the missing words.  We train a Beta regression model\footnote{An alternative to linear regression for cases where the dependent variable is a proportion (0\text{-}1 range).} to predict the concreteness scores of over $22$K words in the historical dataset, 
from the  semantic representations of the words in the $1850$s (again, using embeddings from \citet{hamilton2016diachronic}). The full set of our $600$ declining/stable words was excluded from training, as was a $1000$-word held-out test set. The trained model obtains Pearson's correlation of $0.74$ between scores inferred by our model and the actual ratings for the $1000$-word test set, as well as a correlation of $0.78$ to the ratings of the $461$ rated words in our dataset. Next we use the trained model to predict concreteness rating of all $600$ (declining/stable) words.\footnote{For consistency, we use the predicted scores for all words in our dataset, rather than using the original ratings for those $461$ words that occurred in the \citet{snefjella2019historical} data.}  Among words assigned the highest scores are `verdure' and `diamonds', while their least concrete counterparts include `reasonings' and `magnanimity'. 

\paragraph{Number of Meanings (\texttt{NMngs}).}
\label{sec:historical-meanings}
We make use of the Historical Thesaurus of English (HTE) \citep{kay2019historical}, a database that records the meanings throughout their history for a very large number of words. We are not aware of a resource analogous to HTE for French and German, hence we only consider this factor for English. Each distinct meaning of a word in HTE has recorded its earliest date of use (as well as its latest date of use, for obsolete meanings). We extracted for each word in our English dataset the number of unique meanings it had in $1800$--$1810$.
For example, $10$ distinct meanings were recorded for the word `institution',
but only a single meaning for `ignominious'.  We interpolated the missing values for $168$ words not documented in HTE (split roughly equally between declining and stable words) by assigning to them the mean number of meanings of the $432$ words documented in the database.

\begin{table*}[h!]
\centering
\resizebox{\textwidth}{!}{
\begin{tabular}{lrr|rr|rr}
\toprule
Factor      & severest (D) & longest (S) & solicitude (D) & marriages (S) & ornamented (D) & attracted (S) \\ \hline
\texttt{SemDens}       & \textbf{0.61} & \textbf{0.41} & \textbf{0.50} & \textbf{0.48} & \textbf{0.69} & \textbf{0.51} \\
\texttt{Conc}        & \textbf{0.50} & \textbf{0.77} & \textbf{0.48} & \textbf{0.59} & 0.90 & 0.78 \\
\texttt{NMngs}       & 4.59 & 4.59 & \textbf{2.00} & \textbf{4.59} & 1.00 & 1.00 \\ \hline
\texttt{CDiv}        & \textbf{0.52} & \textbf{0.95} & \textbf{1.31} & \textbf{1.81} & \textbf{1.80} & \textbf{4.40} \\ \hline
\texttt{PhonTyp}     & \textbf{-2.93} & \textbf{-2.24} & \textbf{-3.40} & \textbf{-0.98} & \textbf{-1.62} & \textbf{-1.35} \\
\texttt{PhonDens}    & 6.02 & 5.76 & \textbf{5.87} & \textbf{5.88} & 6.03 & 6.03 \\
\texttt{PhonComp}    & \textbf{0.60} & \textbf{0.40} & \textbf{0.80} & \textbf{0.75} & \textbf{0.40} & \textbf{0.37} \\
\bottomrule
\end{tabular}
}
\caption{Examples of English word-pairs with varying initial frequency, POS, and length, along with their predictor values. `D' indicates a declining word and `S' a stable word. Differences in the expected direction are boldfaced. For convenience, \texttt{CDiv}$\times10^3$ and \texttt{PhonDens}$\times10^{\text{-}3}$ values are presented.}
\label{tbl:predictor-examples}
\end{table*}

\paragraph{Contextual Diversity (\texttt{CDiv}).}
\label{sec:distinctiveness}
For our distributional measure of contextual diversity, we focus on how much the local environment of the target word (i.e., a single word before and after it) deviates from the distribution of words in the language as a whole. For example, consider the words `somewhat' vs.\ `amok' (part of the phrase `run amok'): Because `somewhat' appears in a wide variety of linguistic contexts, the distribution of frequencies of its immediate neighbors will be much more similar to their distribution in the language as a whole, compared to `amok', whose distribution over its neighbors will have a very large peak for the word `run'. \citet{mcdonald2001rethinking} capture this intuition by formulating contextual distinctiveness (the opposite of contextual diversity) as the Kullback-Leibler (KL) divergence between two probability distributions, the conditional distribution of words $c$ in the context of $w$, and the prior distribution of the context words $c$:\footnote{In this study, $c$ ranges over the $10$K most frequent words.  \ssnew{We exclude the top $100$ words as less informative regarding the effect of relative breadth or narrowness of topical distribution on survivability of a word.}}

\vspace{-0.1in}
\begin{equation*}
D_{KL}(P(c|w)||P(c)) = \sum P(c|w)\: \log \dfrac {P(c|w)} {P(c)}
\label{eq:cont-distinctiveness}
\end{equation*}

\noindent
In what follows, we use $D_{KL}(w)$ to mean $D_{KL}(P(c|w)||P(c))$ as defined above, with $c$ understood as our context words.
A higher value for $D_{KL}(w)$ implies that $w$ occurs in a narrow range of contextual usages -- that is, $D_{KL}$ is inversely related to contextual diversity.
To obtain a measure of contextual diversity, we scale $D_{KL}$ to the $0{-}1$ range, by applying a non-linear exponential transformation $1{-}\exp(\text{--}D_{KL})$, and subtract the result from $1$. Formally, contextual diversity of a word $w$ at time period $t$ is defined as:

\vspace{-0.1in}
\begin{equation*}
\texttt{CDiv}^{t}(w){=}\exp(\text{--}D_{KL}^{t}(w))
\label{eq:cont-diversity}
\end{equation*}
\noindent 
Examples of nouns with high contextual diversity in our data are `money', 'effect' and `purchase', while words with low \texttt{CDiv} score include `panegyric', `soldiery' and `rivulet'.

\paragraph{Phonological Typicality (\texttt{PhonTyp}).}

\label{sec:likelihood}
We 
estimate phonological typicality 
using a phoneme-based LSTM \citep{hochreiter1997long} language model\footnote{With two hidden layers (75 and 50 cells), each layer followed by batch-normalization and dropout.}, trained (for each language) on the IPA 
transcriptions \cite{ipa1999handbook} of a $100$K-word sample from the Google ngrams corpus, spanning years $1800$--$1810$, sampled with replacement via multinomial distribution over the word unigram frequencies in the corpus.
Word transcriptions were obtained through
\href{https://github.com/dmort27/epitran}{Epitran} \citep{Mortensen-et-al:2018}, a tool for transcribing orthographic text as IPA, and then manually verified. We chose not to use CELEX \cite{baayen1996celex} (which supports English and German but not French) or a similar lexical resource because Epitran provides broader coverage and manual correction provided acceptable accuracy.
Using the trained language model, the phonological typicality
of a word is the average log probability of the next phoneme conditioned on the word's prefix. 

Formally, for a word $w$ with length $k$:

\vspace{-0.10in}
\begin{equation*}
\texttt{PhonTyp}(w){=}\dfrac {\sum \log P(c_i\:|\:c_1, .., c_{i-1})} {k}, i{\in}{[1..k]}
\label{eq:likelihood}
\end{equation*}



\paragraph{Phonological Density (\texttt{PhonDens}).}
\label{sec:phon-density}
Following 
\citet{bailey2001determinants}, we computed phonological density of a word as the sum of distances of its IPA transcription to that of all other word types comprising the lexicon
in $1800$--$1810$.
Formally, phonological density of a word $w$ with respect to a lexicon $L$ is defined as:

\vspace{-0.15in}
\begin{equation*}
\texttt{PhonDens}(w) = \sum_{v \in L} \exp(\text{--}d(w,v))
\label{eq:phon-density}
\end{equation*}

\noindent 
where the distance $d$ is the normalized Levenshtein distance \citep{levenshtein1966binary} between the phonetic forms of words $w$ and $v$.

\paragraph{Phonological Complexity (\texttt{PhonComp}).}
\label{sec:complexity}

Words can be phonologically complex in various dimensions.
For ease of calculation across the three languages in this study, we measured one of these, the ratio of syllables to segments, by counting the number of syllabic nuclei (vowels) and the number of phonemes (segments). Vowels and segments in aforementioned IPA transcriptions were classified as such according to the specifications given by the International Phonetic Association. A higher ratio was taken to indicate greater phonological complexity, corresponding to greater `syllable density'.


\paragraph{Examples of Word Pairs and Factor Values.}

Table~\ref{tbl:predictor-examples} presents three examples of English word-pairs along with the values computed for these $7$ factors. The vast majority of differences occur in the predicted direction, with a few exceptions (e.g., the higher degree of concreteness of the declining `ornamented' vs.\ the stable `attracted'). All three declining words exhibit notably higher \texttt{SemDens}, lower \texttt{CDiv} (extremely so for `ornamented'), lower \texttt{PhonTyp} (extremely so for `solicitude'), and higher \texttt{PhonComp}.


\section{Results and Discussion}
\label{sec:results}

\begin{table*}[h]
\centering
\resizebox{\textwidth}{!}{
\begin{tabular}{lrr|rr|rr}
\toprule
& \multicolumn{2}{c|}{\textbf{English}} & \multicolumn{2}{c|}{\textbf{French}} & \multicolumn{2}{c}{\textbf{German}} \\
Factor & \multicolumn{1}{c}{dec} & \multicolumn{1}{c|}{stb} & \multicolumn{1}{c}{dec} & \multicolumn{1}{c|}{stb} & \multicolumn{1}{c}{dec} & \multicolumn{1}{c}{stb}  \\ \hline
\texttt{SemDens} & 0.55** ($\pm{0.07}$) & 0.52 ($\pm{0.07}$) & 0.65** ($\pm{0.10}$) & 0.53 ($\pm{0.07}$) & N/A & N/A \\
\texttt{Conc} & 0.53*\hspace{0.5em} ($\pm{0.15}$) & 0.57 ($\pm{0.16}$) & N/A & N/A & N/A & N/A \\
\texttt{NMngs} & 3.91** ($\pm{2.21}$) & 5.26 ($\pm{4.02}$) & N/A & N/A & N/A & N/A \\ \hline
\texttt{CDiv} & 1.97** ($\pm{4.10}$) & 2.93 ($\pm{7.72}$) & 0.88** ($\pm{2.82}$) & 1.20 ($\pm{3.30}$) & 1.47** ($\pm{2.01}$) & 2.01 ($\pm{4.05}$) \\ \hline
\texttt{PhonTyp} & -2.02*\hspace{0.5em} ($\pm{0.85}$) & -1.85 ($\pm{0.71}$) & -2.27** ($\pm{0.84}$) & -2.00 ($\pm{0.86}$) & -1.83** ($\pm{0.47}$) & -1.73 ($\pm{0.46}$) \\
\texttt{PhonDens} & 5.90\hspace{1em} ($\pm{0.12}$) & 5.92 ($\pm{0.12}$) & 5.37\hspace{1em} ($\pm{0.11}$) & 5.38 ($\pm{0.12}$) & 8.65\hspace{1em} ($\pm{0.27}$) & 8.65 ($\pm{0.26}$) \\
\texttt{PhonComp} & 0.38*\hspace{0.5em} ($\pm{0.07}$) & 0.35 ($\pm{0.07}$) & 0.38\hspace{1em} ($\pm{0.10}$) & 0.37 ($\pm{0.09}$) & 0.45\hspace{1em} ($\pm{0.09}$) & 0.44 ($\pm{0.09}$) \\
\bottomrule
\end{tabular}
}
\caption{Mean ($\pm$SD) of factor values for declining (dec) and stable (stb) words.
Significant differences are marked by `**' ($p{\textless}.001$) and `*' ($p{\textless}.01$). 
For convenience, \texttt{CDiv}$\times10^3$ and \texttt{PhonDens}$\times10^{{\text{-}3}}$ values are presented. All significant differences in factors match the direction of our prediction in Table~\ref{tab:factors-and-predictions}.
}
\label{tbl:predictor-values}
\end{table*}

\subsection{Factor Analysis}
\label{sec:factor-analysis}
We aim to test the predictive power of our $7$ factors 
on a word's likelihood to fall out of use.
As a first step,
we assess the difference in the defined predictors across the two sets of declining and stable words in each language, by applying statistical significance tests on individual factor values. Specifically, we apply the Wilcoxon pairwise sign-ranked test on the values for each predictor, testing whether the two (paired) samples exhibit a significant difference in each case. Table~\ref{tbl:predictor-values} reports the results for the three languages, split by factor categories --- semantic, distributional and phonological. 
All our predictions (see Table~\ref{tab:factors-and-predictions}) are borne out, except for \texttt{PhonComp} (with a significant difference only for English) and \texttt{PhonDens} (insignificant for all languages).

\begin{figure*}[hbt]
\centering
\includegraphics[width=15.3cm]{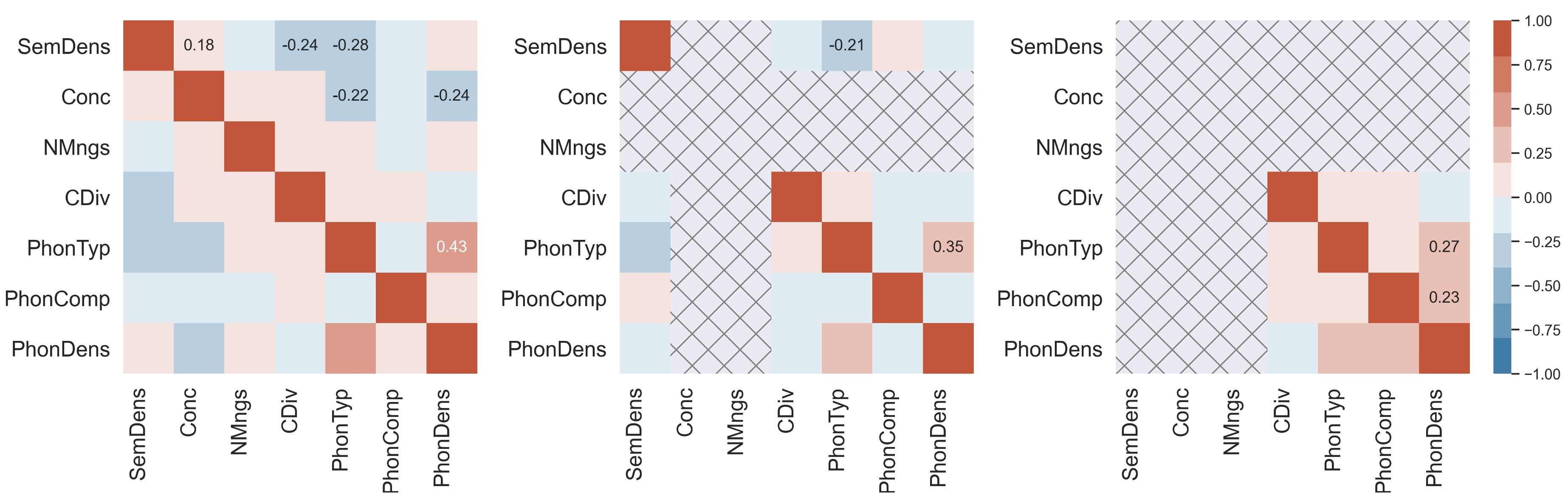}
\vspace{-0.1in}
\caption{Heatmap of correlations of predictors for English (left), French (middle), and German (right). 
Uncolored rows/columns denote unavailable measures in French and German. 
Numeric values are shown only for significant correlations (after applying Bonferroni correction for multiple comparisons). 
}
\label{fig:factor-correlation}
\end{figure*}

Figure~\ref{fig:factor-correlation} presents the Pearson correlations between the predictors as a heatmap (predictors missing from French and German are left uncolored). 
There is only one moderate correlation, of \texttt{PhonTyp} 
with \texttt{PhonDens}, which is attributable
to the fact that atypically pronounced words will tend to have fewer close phonological neighbors, and thus sparser phonological neighborhoods.

\subsection{\ssnew{Predicting a Word's Future Status}}
\label{sec:predict-regression}

Here we test whether the systematic and significant differences among our $7$ factors, as observed in Table~\ref{tbl:predictor-values}, support their use in a prediction task regarding lexical decline.  Because each declining word in our data is matched to a (control) stable word, we use a logistic regression model to predict the future status of the words in each pair: which is the declining word, and which the stable one.
We examine individual regressor coefficients to assess the relative contribution of individual features to the prediction task.
\tochecknew{We also report the pseudo-$r^2$ of the regression model, as an indication of the collective predictive power of our factors.\footnote{We report here the results of a logistic regression model, \tochecknew{using the Python GLM Logit implementation from \url{https://www.statsmodels.org}, with the pseudo{-}$r^2$ calculation provided at \url{https://www.statsmodels.org/devel/discretemod.html}.}  In Appendix A.4, we provide the (complementary) results of a logistic regression-based classification task.}}

Specifically, each item in this task is a word-pair from our matched sets of declining (\texttt{dec}) and stable (\texttt{stb}) words (e.g., `thence':`forward'), where the items are created such that (a random) half of the pairs are in the order \texttt{dec:stb} and the other half are in the order \texttt{stb:dec}.  \ssnewer{The dependent variable in the logistic regression is a binary variable indicating whether the item is in the order \texttt{dec:stb} (a value of $1$) or \texttt{stb:dec} (a value of $0$).}  The $7$ independent variables in the regression are formed by taking the difference between the corresponding feature values of each word in the pair (all features scaled to the $0$--$1$ range).
As an example, for the `thence':`forward' word-pair, the $7$ predictors are calculated by subtracting the values of each of the $7$ features of `forward' from those of `thence', and the dependent variable is defined as `1', for \texttt{dec:stb}. 
We run a regression of this form on each of the three languages; we present detailed results on English, with comparison to French and German.\footnote{We also ran a model adding features for the differences in frequency and length for each matched word-pair; as expected, the results were unaffected, confirming the quality of matching on these covariates.}

In English, the logistic regression obtained a pseudo{-}$r^2$ of $0.23$, while a similar analysis for French achieved a pseudo{-}$r^2$ of $0.41$.\footnote{The higher value for French seems due to a number of declining scientific terms distinguished by a much higher average \texttt{SemDens}, compared to stable words.}  A pseudo{-}$r^2$ of only $0.08$ was obtained for German, which has no semantic features available.
%
Table~\ref{tbl:diff-regression-eng} presents the detailed results of the model for English.  All of the semantic features (\texttt{SemDens}, \texttt{Conc}, \texttt{NMngs}) and the single distributional feature (\texttt{CDiv}) have a significant contribution to the model.  Moreover, the sign of the $\beta$ coefficient in each case matches the direction of effect that we hypothesized, in line with the individual factor analysis in \Section{sec:factor-analysis} above.
(For example, a positive difference in \texttt{SemDens} is indicative of a \texttt{dec:stb} word-pair, annotated with the label `1' in our analysis, because \texttt{SemDens} values of declining words tend to be higher.)  On the other hand, none of the phonological features contribute to the model. 
The results on French showed a similar pattern: \texttt{SemDens} was strongly predictive of decline, while \texttt{CDiv} was marginally so.  In German, \texttt{CDiv} was significantly predictive, as was \texttt{PhonTyp}; it isn't clear whether phonological form is actually more important in German, or is simply seen to play a role when no semantic features are available.  

We conclude that semantic and distributional features may be associated with aspects of lexical access and learning that are strong enough to influence word choice and consequent trends in frequency, while phonological effects may only ``fine-tune'' word preferences that are largely shaped by semantic need.

\begin{table}[h]
\centering
\resizebox{\columnwidth}{!}{
\begin{tabular}{lrrrr}
\toprule
predictor & \multicolumn{1}{c}{$\beta$ coeff.} & \multicolumn{1}{c}{std err($\beta$)} & \multicolumn{1}{c}{$z$} & \multicolumn{1}{c}{$p$}  \\ \hline
const       & 0.018     & 0.135     & 0.137     & 0.891 \\ \hline
\textbf{\texttt{SemDens}}     & 0.589     & 0.154     & 3.825     & 0.000 \\
\textbf{\texttt{Conc}}        & -0.513    & 0.150     & -3.426    & 0.001 \\
\textbf{\texttt{NMngs}}       & -0.847    & 0.204     & -4.147    & 0.000 \\ \hline
\textbf{\texttt{CDiv}}        & -1.491    & 0.472     & -3.176    & 0.002 \\ \hline
\texttt{PhonTyp}     & -0.262    & 0.158     & -1.661    & 0.097 \\
\texttt{PhonDens}    & -0.052    & 0.150     & -0.350    & 0.726 \\
\texttt{PhonComp}    & 0.218     & 0.149     & 1.466     & 0.143 \\

\bottomrule
\end{tabular}
}
\caption{Logistic regression analysis predicting word-pair direction (1: \texttt{dec:stb}, or 0: \texttt{stb:dec}) from pairwise differences in factor values. Significant predictors in bold.
}
\label{tbl:diff-regression-eng}
\end{table}

\section{Diachronic Analysis of Lexical Loss}
\label{sec:diachronic-analysis}
We next explore whether there are diachronic patterns in the contextual dissemination of words, over the $21$ decades of our data, that differ between declining and stable words.
A specific question is whether a word falling out of use in a language uniformly reduces its frequency across the entire diversity of its contextual environments, or if it instead gradually `abandons' particular contextual niches, thereby narrowing its linguistic dissemination.
We hypothesize that declining and stable words differ in the diachronic trend of their \texttt{CDiv} values; specifically, that declining words gradually fade out from certain contextual usages, thereby reducing the number of linguistic environments they populate \cite{traugott2001regularity}. To corroborate this, we perform diachronic analysis of contextual diversity.
\ellanew{We approach this question by using linear regression to fit a temporal trend line over each word $w$'s \texttt{CDiv} values, across the $21$ decades -- i.e., regressing \texttt{CDiv}$_t^w$ on $t{\in}[1..21]$. We expect this trend line to show a decreasing tendency for declining words, indicative of contextual shrinkage, and a stable or increasing tendency for stable words, indicative of stability or growth of contexts. In particular, the regression line coefficients of the declining set should be significantly lower than that of stable words.}

However, we must adopt a multiple regression approach that incorporates variables (other than time) that could also contribute to variation in a word's \texttt{CDiv} values. 
Specifically, we identified two properties that may bias the \texttt{CDiv} of a word when comparing across decades:\footnote{When using \texttt{CDiv} in Section~\ref{sec:results}, this was not an issue, since we restricted the focus to a single decade, $1800$--$1810$.} (1)~the number of unique books used for data extraction in that decade, and (2)~the frequency of the word in that decade. First, a greater number of unique per-decade books is likely to increase contextual diversity, since a higher number of distinct literature sources raises the chance of a wider range of contextual domains. Second, lower frequency of a word is likely to negatively affect its contextual diversity -- the lower the frequency, the less opportunity there is for a word to occur in different contexts.
We address these potential confounds by using the per-decade values of each of these properties as additional independent variables, along with time $t$, in a multiple regression.\footnote{We compute per-decade number of books by summing the number of unique books reported in the Google-ngrams dataset for all years of the decade. We then take the $log$ of this value since the relative increase in \texttt{CDiv} due to number of books is likely attentuated as this number grows, motivating the use of a sub-linear function.}

\ellanew{This yields the following regression model:} 

\vspace{-0.14in}
\begin{equation*}
\resizebox{\columnwidth}{!}{
$\texttt{CDiv}_t^w = \beta_0 + \beta_1 * log(B_t) + \beta_2 * F_t^w + \beta_3 * t + \epsilon_t^w$
}
\label{eq:diachronic_cdiv}
\end{equation*}

\noindent
\ellanew{where $\beta_3$, the coefficient of the decade counter $t{\in}[1..21]$, reflects the temporal trend of \texttt{CDiv} for each word $w$ in our data: the sequential tendency of $w$'s contextual diversity over time, taking into account the effects of number of books, $log(B_t)$, and word frequency, $F_t^w$, in each decade $t$.}

\ellanew{Our analysis now proceeds by assessing the distribution of the $\beta_3$ coefficients.  As noted above, we hypothesize that these coefficients will differ across declining and stable words; specifically, declining words will tend to have negative $\beta_3$ coefficients, indicating decreasing contextual diversity over time, while stable words will have non-negative $\beta_3$ coefficients, showing a flat or increasing tendency of diversity.}

\begin{figure}[hbt]
\centering
\resizebox{\columnwidth}{!}{
\includegraphics{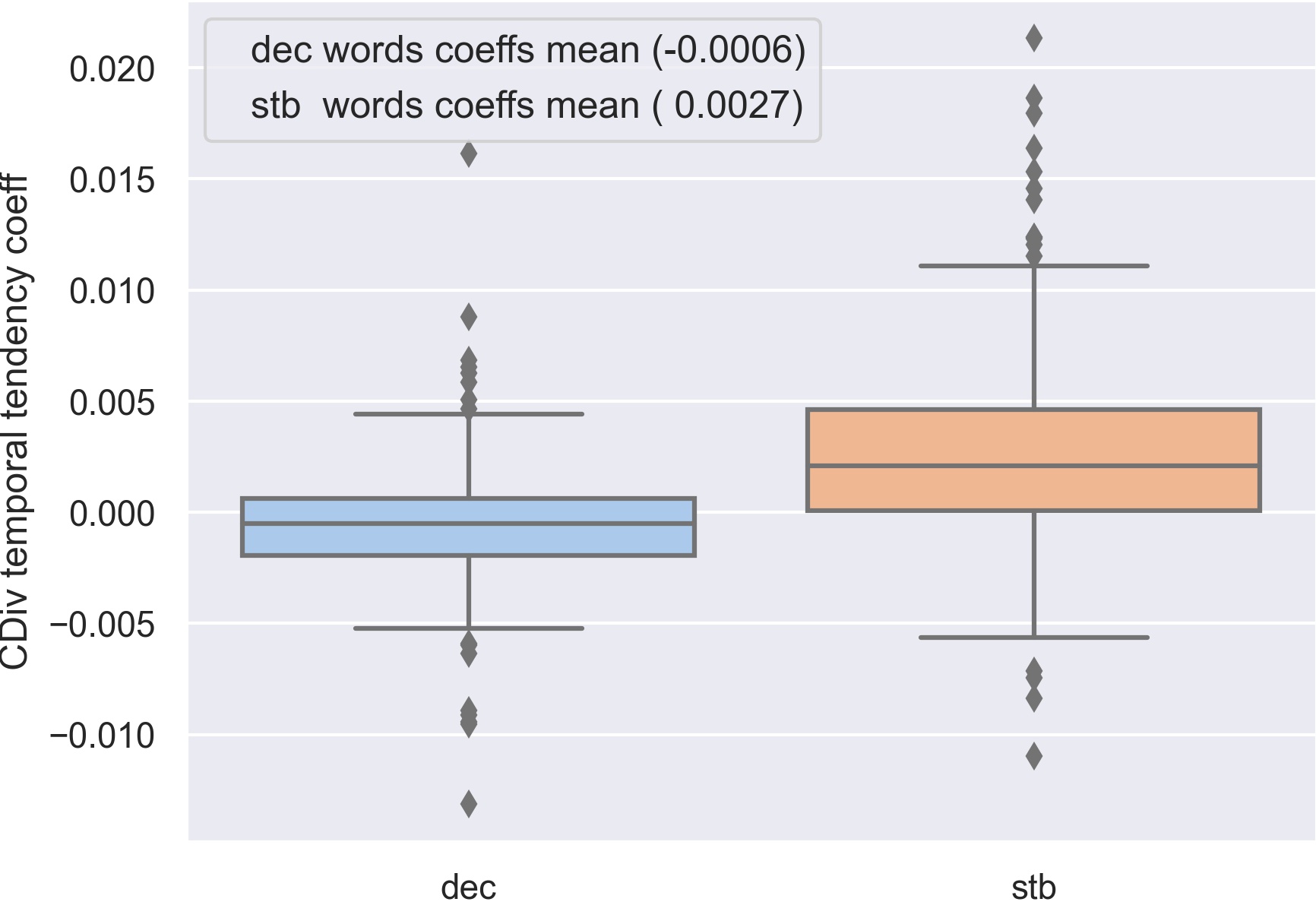}
}
\vspace{-0.15in}
\caption{Boxplot of the distribution of the two sets of $\beta_3$ coefficients: for declining (left) and stable (right) words in our English dataset.}
\label{fig:cdiv-tendency}
\end{figure}

\ellanew{Figure~\ref{fig:cdiv-tendency} presents two boxplots of the distributions of the two sets of $\beta_3$ coefficients -- for the $300$ declining words (left) and  $300$ stable words (right) in our English dataset.\footnote{Similar results were found for French and German.}}
\ssnewer{The means of the two distributions significantly differ from each other, as well as from $0$, when applying a Wilcoxon test ($p${\textless}$0.001$ for all tests).}
\ellanew{We thus find support for the claim that declining and stable words have different diachronic patterns of contextual diversity.
The negative mean and median of the coefficients for the declining words (mean{=}$-0.0006$; median{=}$-0.0005$) further support our specific hypothesis of diachronic contextual loss for these words. 
In contrast, the coefficients of stable words have a positive mean and median (mean{=}$0.0027$; median{=}$0.0021$). Although the mean coefficient values are small, the coefficients for the stable words are consistently larger, as quantified by the Wilcoxon test.
Moreover, the wide range of their (mostly positive) coefficients indicates a strong tendency of stable words to increase in contextual diversity and gradually occupy a broader range of environments, contrasting with declining words.}

\begin{table*}[h!]
\centering
\resizebox{0.85\textwidth}{!}{
\begin{tabular}{ll|ll|ll}
\toprule
\multicolumn{2}{c|}{\textbf{English}} & \multicolumn{2}{c|}{\textbf{French}} & \multicolumn{2}{c}{\textbf{German}} \\ \hline
\multicolumn{1}{c}{dec} & \multicolumn{1}{c|}{stb} & \multicolumn{1}{c}{dec} & \multicolumn{1}{c|}{stb} & \multicolumn{1}{c}{dec} & \multicolumn{1}{c}{stb} \\ \hline
verdure & criminals & industrieux & législative & tugendhaft & schwarzer \\
impracticable & unreasonable & évacuations & inventions & dünkt & hängen \\
unexampled & invaluable & estimable & acquises & endigen & brauche \\
dignities & extinction & intrépidité & irrégularité & hernach & innen \\
insensibility & embarrassment & factieux & habituel & mannigfaltige & gegenseitigen \\
amusements & foundations & mâchoire & surprise & füglich & dringend \\
illustrious & successful & magnésie & désert & siebenten & tägliche \\
necessaries & repetition & réfraction & conversion & redlichen & einseitigen \\
sublimity & attainment & sulfurique & naturelles & erstlich & einziges \\
whence & highly & prairial & arbitraire & dermalen & halbes \\

\bottomrule
\end{tabular}
}
\vspace{-0.1cm}
\caption{Examples of declining--stable word pairs for English, French and German, selected according to the policy described in Section~\ref{sec:word-sampling}, further detailed in Appendix A.}
\label{tbl:word-pair-examples}
\end{table*}

\section{Conclusions}
\label{sec:conclusions}

We have proposed factors of various types --- semantic, distributional, and phonological --- and shown that the semantic and distributional features are robust predictors of whether a word will remain stable in frequency or fall into decline. In particular, we have focused on factors that can influence the cognitive processing of words, affecting how likely they are to be used and learned.  
Given that broad external influences, such as language contact, as well as social and technological developments, are known to have a massive effect on the content of vocabulary, this study constitutes an important demonstration of the potential influence of internal cognitive mechanisms on the `survivability' of words. Our findings suggest that factors affecting a word's trajectory are more likely to be semantic or distributional than phonological, perhaps because speakers or writers, when they are looking for a word, are guided primarily by syntactic and semantic criteria.


The behavior of most of the factors we proposed matches the expectations for declining vs.\ stable words that were motivated by the psycholinguistic literature.  Our findings are consistent with an evolutionary view where psycholinguistic factors influence the `reproductive fitness' --- the fitness for self-perpetuation --- of words. This, in turn, supports a broader evolutionary research agenda in historical linguistics.


\section*{Appendix A}
\label{sec:appendix-a}

\paragraph{A.1: Word frequency normalization.} 
As noted in \Section{sec:word-sampling}, we normalize the frequencies of each word across all decades before calculating the fit (in MSE) of its frequency curve to the target `decline' or `stable' curve.  First, due to the widely varying amounts of data available in each decade, we take the frequency of a word as its relative frequency within each decade.  Further normalization was motivated by our observation that words at different frequency levels could have very different MSE values w.r.t.\ the fit curves, with higher frequencies generally leading to higher MSEs. 

\begin{figure}[h!]
\centering
\resizebox{\columnwidth}{!}{
\includegraphics{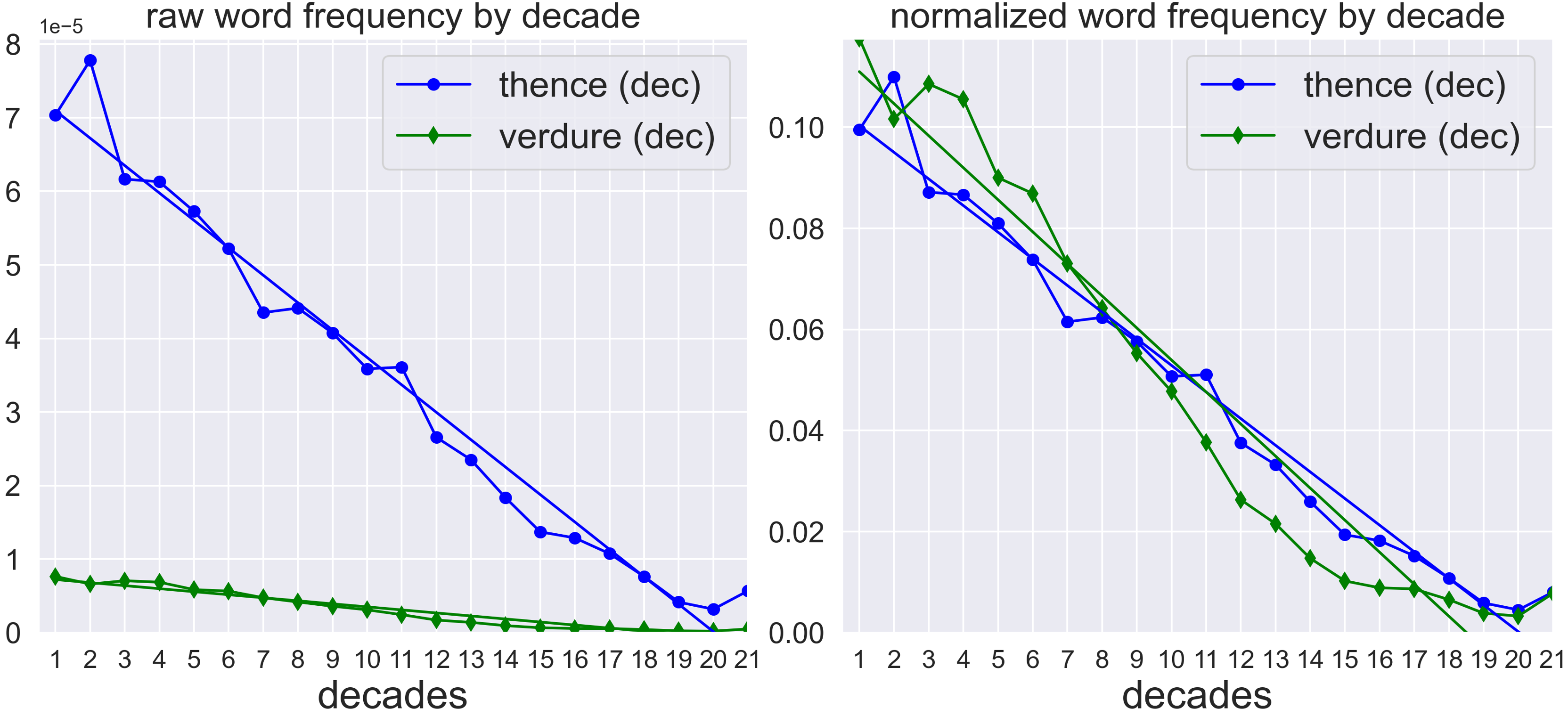}
}
\caption{Per-decade frequencies of two declining words and their corresponding fit lines: raw frequencies (left) and normalized frequencies (right).}
\label{fig:regression-fit}
\end{figure}

An example is illustrated in the left panel of \Figure{fig:regression-fit}, which shows the per-decade relative frequencies of two declining words -- `thence' and `verdure' -- with their corresponding (declining) fit line.  (Recall we take the MSE between a piecewise curve with a declining piece and a `zero' piece that is horizontal at $0$; the lines shown in \Figure{fig:regression-fit} are the declining pieces.)  The higher initial frequency of `thence' potentially contributes to higher MSE: a $10\%$ offset from the fit line contributes more to MSE of `thence' than of `verdure'.  Normalization of each word's frequencies (dividing by its total frequency across the decades) eliminates this confound by yielding a curve that reflects relative change across the decades, as exemplified in the right panel of \Figure{fig:regression-fit}.

\vspace{.1in}
\noindent
\textbf{A.2: Finding matched stable words.} The matching procedure greedily matches each declining word with the first stable counterpart that meets all three constraints (on initial frequency, length, and POS) by traversing the list of stable words sorted by their `stability' measure, so as to exploit the most stable words first.  The matched stable word is then removed from the stable list, so that it will not be considered for further matches.

The Wilcoxon pairwise sign-ranked test on the frequency and length of the matched word-pairs revealed no significant differences, implying that no bias was introduced into the selection process with respect to the control factors.

\label{sec:appendix-b}

\vspace{.1in}
\noindent
\textbf{A.3: Example declining--stable pairs.} Table~\ref{tbl:word-pair-examples} presents $10$  sample word-pairs for English, French and German. Recall that words are matched by initial frequency ($\pm10\%$), length in characters ($\pm2$ characters, with the additional restriction that the sum of lengths of all stable words must be within $1$ of the sum of lengths of all declining words), and POS (i.e., nouns were matched with nouns, adjectives with adjectives, etc.).


\vspace{.1in}
\noindent
\ellanew{\textbf{A.4: Predicting a word's future status in a classification task.}
The regression analysis in Section~\ref{sec:predict-regression} can alternatively be formulated as a classification task distinguishing the declining and stable word of a pair.  
Each classification item is a word-pair from our matched sets of declining (\texttt{dec}) and stable (\texttt{stb}) words (e.g., `thence':`forward'), concatenating the $n$ features extracted for each of the two words into a single feature vector of $2n$ values, where the first half represents the first word in the pair (e.g., `thence') and the second half, the second word (e.g., `forward').  The items are created such that (a random) half of the pairs are in the order \texttt{dec:stb} and the other half \texttt{stb:dec}, with the appropriate training label; for a test item, the classifier must output \texttt{dec:stb} or \texttt{stb:dec}.  Due to the relatively small dataset, we use a leave-one-out evaluation paradigm.  Average classification accuracy higher than random ($0.5$) will be indicative of the predictive power of our identified factors.}

\ellanew{Using the classifier version of logistic regression,\footnote{\url{https://scikit-learn.org/stable/modules/generated/sklearn.linear_model.LogisticRegression.html}} we obtain a classification accuracy of $0.67$, $0.80$, and $0.61$ for English, French, and German,
respectively.  }
\tocheck{Recall that French has many declining medical terms with a higher \texttt{SemDens}, leading to an easier classification task, while German has no semantic features available, which were shown to be highly predictive of decline in the other languages.  Although the accuracy for English is not high, it is well above random, and it must be remembered that we are only testing our cognitive features, and not including the myriad social and cultural influences on lexical change.}

\ellanew{The results here further support our findings in Section~\ref{sec:predict-regression}, indicating again that the features we have proposed have useful predictive power in identifying the declining word of a pair that shares similar frequency, length, and POS.}

\section*{Acknowledgements}
We are grateful to the Action Editor, Jacob Eisenstein, and the anonymous reviewers for their constructive and detailed feedback which helped us improve the research. We are also thankful to Yang Xu from the Language, Cognition, and Computation (LCC) Group at the University of Toronto for offering comments on an earlier version of this work. 
This research was supported by NSERC grant RGPIN-2017-06506 to Suzanne Stevenson. 
This material is based in part on research sponsored by the Air Force Research Laboratory under agreement number FA8750-19-2-0200. The U.S. Government is authorized to reproduce and distribute reprints for Governmental purposes notwithstanding any copyright notation thereon. The views and  conclusions contained herein are those of the authors and should not be interpreted as necessarily representing the official policies or endorsements, either expressed or implied, of the Air Force Research Laboratory or the U.S. Government.

\bibliographystyle{acl_natbib}
\bibliography{lexical-decline}

\end{document}
